\documentclass{article}
\usepackage{spconf,amsmath,graphicx}
\usepackage{amssymb,amsfonts}
\usepackage{graphicx}
\usepackage{textcomp}
\usepackage{booktabs}
\usepackage{tabularx}
\usepackage[dvipsnames]{xcolor}
\usepackage{soul}
\usepackage{lipsum}
\usepackage{inconsolata}
\usepackage{tipa}
\usepackage{url}
\usepackage{float}
\usepackage{algorithm}
\usepackage{algpseudocode}
\usepackage{enumitem}
\usepackage[
 bibencoding=utf8,
 backend=biber,
 bibstyle=ieee,
 citestyle=numeric-comp,
 defernumbers=true,
 sortcites=true,
 maxbibnames=20,
 minbibnames=4,
]{biblatex}
\usepackage[export]{adjustbox}
\usepackage[caption=false, font=footnotesize]{subfig}
\usepackage{hyperref}
\hypersetup{
    colorlinks=true,
    linkcolor=blue,
    urlcolor=blue,
    citecolor=blue,
}
\usepackage[nameinlink, capitalise]{cleveref}
\usepackage{pifont}
\def\BibTeX{{\rm B\kern-.05em{\sc i\kern-.025em b}\kern-.08em
    T\kern-.1667em\lower.7ex\hbox{E}\kern-.125emX}}

\title{Exploring the Impact of Data Quantity on ASR in Extremely Low-resource Languages}
\name{Yao-Fei Cheng$^{1}$, Li-Wei Chen$^{2,3}$, Hung-Shin Lee$^{3}$, Hsin-Min Wang$^{4}$}
\address{$^{1}$University of Washington, $^{2}$National Tsing Hua University\\$^{3}$United Link Co., Ltd., $^4$Academia Sinica\\nlp5566@uw.edu}
\begin{document}
%
\maketitle

\begin{abstract}
This study investigates the efficacy of data augmentation techniques for low-resource automatic speech recognition (ASR), focusing on two endangered Austronesian languages, Amis and Seediq. Recognizing the potential of self-supervised learning (SSL) in low-resource settings, we explore the impact of data volume on the continued pre-training of SSL models. We propose a novel data-selection scheme leveraging a multilingual corpus to augment the limited target language data. This scheme utilizes a language classifier to extract utterance embeddings and employs one-class classifiers to identify utterances phonetically and phonologically proximate to the target languages. Utterances are ranked and selected based on their decision scores, ensuring the inclusion of highly relevant data in the SSL-ASR pipeline. Our experimental results demonstrate the effectiveness of this approach, yielding substantial improvements in ASR performance for both Amis and Seediq. These findings underscore the feasibility and promise of data augmentation through cross-lingual transfer learning for low-resource language ASR.
\end{abstract}


\section{Introduction}
\label{sec:intro}



The development of contemporary end-to-end automatic speech recognition (ASR) systems has achieved remarkable success in high-resource languages such as English, French, and Chinese, largely due to the availability of extensive speech-text paired data \cite{gulati2020,kim2022}. However, this abundance does not extend to endangered languages like Amis and Seediq, where acquiring high-quality transcriptions is particularly challenging and resource-intensive. The scarcity of annotated data, coupled with factors such as limited speaker populations and non-standardized writing systems, poses significant obstacles to building effective ASR systems. In response, self-supervised learning (SSL) has emerged as a promising approach to mitigate data sparsity in low-resource ASR settings \cite{schneider2019,baevski2020,hsu2021a}. SSL models, pre-trained on large amounts of unlabeled speech, can be fine-tuned with limited labeled data. However, the pre-training stage typically requires far more data than supervised methods—for instance, the \texttt{wav2vec 2.0 base} model demands 960 hours of speech for effective pre-training \cite{baevski2020}. This makes training SSL models from scratch impractical for endangered languages, where large-scale data collection is often infeasible.

To address the broader issue of resource scarcity in natural language processing, researchers have explored various strategies, including language representation tools like \texttt{lang2vec} \cite{littell2017} and data augmentation techniques \cite{bartelds2023,safri2023}. However, these approaches face limitations when applied to low-resourced languages like Amis and Seediq. \texttt{lang2vec} currently lacks support for these languages, while data augmentation often depends on textual data, which is scarce or unavailable for many spoken and low-resource languages.

Consequently, recent research has shifted towards leveraging multilingual SSL models for low-resource ASR \cite{conneau2021,wang2021}. This approach capitalizes on phonetic similarities across languages, enabling knowledge transfer and improved transcription accuracy during fine-tuning. Further emphasizing the potential of this approach, Nowakowski et al. demonstrated the effectiveness of continuing pre-training with target language data, albeit with a significant data requirement exceeding $234$ hours \cite{nowakowski2023}. Their work also explored leveraging phonetically similar languages for data augmentation, achieving promising results. Similarly, San et al. proposed utilizing donor speech data from related languages for fine-tuning, highlighting the potential of cross-lingual transfer learning in low-resource settings \cite{san2024}.

This research diverges from previous studies by exploring the feasibility of achieving robust ASR performance with minimal paired and unpaired data for the target language. Focusing on the severely low-resource Austronesian languages, Amis and Seediq, we aim to develop effective ASR systems using a modest multilingual corpus and limited paired data (less than one hour per language). Our approach centers on a novel data-selection strategy designed to identify and incorporate utterances from the multilingual corpus that exhibit close phonetic and phonological similarity to the target languages, thus refining the SSL-ASR pipeline. We achieve this by first employing a language classifier to extract language-specific embeddings for each utterance \cite{jia2022,jia2023}. Then, leveraging a suite of one-class classifiers – One-class SVM \cite{scholkopf2001}, Isolation Forest \cite{liu2008}, and Deep SVDD \cite{ruff2018} – we rank and select utterances based on their decision scores, prioritizing those most similar to the target languages. Our results demonstrate the effectiveness of this approach for both Amis and Seediq, yielding promising ASR performance despite the significant data constraints.


\section{Data Description}
\label{sec:data}

\begin{figure}[t]
\centering
\includegraphics[width=0.45\textwidth]{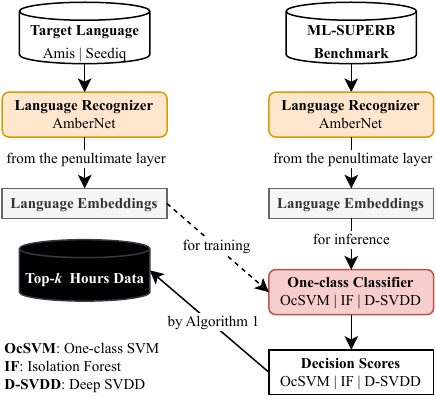}
\caption{Diagram for picking the top-$k$ hours data.}
\label{fig:one_class}
\vspace{-15pt}
\end{figure}

Taiwan is home to $16$ indigenous ethnic groups, encompassing $42$ dialects. The Indigenous Languages Research and Development Foundation (ILRDF) introduced Klokah\footnote{\url{https://web.klokah.tw}}, an e-learning platform with abundant audio materials to foster native language use among the younger generation. For ASR system development, we curated a multi-speaker dataset from two subjects: reading-and-writing and contextual-language, chosen based on their suitable audio lengths. We selected Amis and Seediq for experimentation, allocating 1 hour or 10 minutes of speech for training and 10 minutes each for validation and testing.



\subsection{Amis}
\label{ssec:amis}

Amis has approximately $218,500$ speakers in Taiwan. It is one of the endangered languages identified by the United Nations Educational, Scientific, and Cultural Organization (UNESCO). Its writing system consists of $21$ Latin letters and one digraph (ng). Another $5$ Latin letters (b, g, j, q, and v) are used only for loanwords. The epiglottal stop /\textipa{\textbarglotstop}/ and glottal stop /\textipa{\textglotstop}/ are marked as \textit{’} and \textasciicircum, respectively. The vowel length /\textipa{\textlengthmark}/ is marked as a colon symbol $\colon$, meaning semantic emphasis.

Amis has five dialect groups: 'Amisay a Pangcah, Siwkolan 'Amis, Pasawalian Pangcah, Farangaw Amis, and Palidaw 'Amis. Its written system can be vague, i.e., the same pronunciation has different written symbols. For instance, only 'Amisay a Pangcah uses the symbol \textit{o} to represent /\textipa{o}/, and the other four groups use the symbol \textit{u} to represent /\textipa{o}/.

\subsection{Seediq}
\label{ssec:seediq}

Seediq is also an endangered language identified by UNESCO. Its language population in Taiwan is about $10,970$. The written system of Seediq consists of $23$ Latin letters and seven diphthongs (aw, ay, ey, ow, uy, ug, and ng). The Seediq text is written in syllabic units of semantic words, with an apostrophe \textit{'} to mark where the syllables in the same word have to be separated because of the sound level or where there is still a problem with the attribution of the pronunciation, or where the vowel sound /\textipa{\textglotstop}/ is indicated.




\section{Method}
\label{sec:research}

Inspired by Nowakowski et al.'s work \cite{nowakowski2023}, which demonstrated that continued pre-training with a related language (Japanese) significantly improved ASR performance for Ainu, a low-resource language\footnote{Interestingly, the Ainu speech data utilized for continued pre-training of the SSL model exceeded 200 hours.}, we propose a novel data augmentation strategy. Since \texttt{lang2vec} \cite{littell2017}—the tool they used to identify related languages—does not support our target languages, we present an alternative approach: leveraging multilingual fine-tuning of an SSL model with data from languages that share phonetic and phonological similarities with our target languages. This strategy aims to enhance speech recognition performance by enriching the training data with acoustically similar utterances from related languages.

Our methodology, illustrated in \cref{fig:one_class}, employs a two-phase approach. In the first phase, a neural spoken language identification model analyzes utterances to extract language-specific embeddings \cite{jia2022,jia2023} (orange box in \cref{fig:one_class}). These embeddings, derived from the penultimate layer of the classifier, encapsulate rich phonetic and phonological information. The second phase leverages these embeddings to train a suite of one-class classifiers (red box in \cref{fig:one_class}). These classifiers are then employed to identify and select utterances from a larger multilingual dataset that exhibit high similarity to the target language, thereby enriching the training data for subsequent SSL model fine-tuning.

\begin{algorithm}[t]
\caption{Multi-list Utterance Selection}
\label{alg:extract}
\begin{algorithmic}[1]
\small
\State \textbf{Given} three lists of utterances $U_1$, $U_2$, and $U_3$ sorted by decision scores of Deep SVDD, One-class SVM, and Isolation Forest \\
Initial ranking limit $L_0$, final list $R$, and specified hours $k$.
\State $L \gets L_0$
\While {the total length\ in $R < k$ hours}
\State $\widehat{U}_1 \gets U_1[:L]$
\State $\widehat{U}_2 \gets U_2[:L]$
\State $\widehat{U}_3 \gets U_3[:L]$
\For{\text{$u \in \widehat{U}_1 $}}
\If {$u \notin R$ \textbf{and} $u \in \widehat{U}_2 $ \textbf{and} $u \in \widehat{U}_3 $}
\State $R$.append($u$)
\EndIf
\EndFor
\State $L \gets L + L_0$
\EndWhile
\end{algorithmic}
\end{algorithm}




First, One-class Support Vector Machine (OcSVM) aims to construct a hyperplane in a high-dimensional feature space that encloses the majority of data points while maximizing the distance between this hyperplane and the origin \cite{scholkopf2001}. This approach effectively separates the data from regions with a sparse data density. Data points falling outside the hyperplane's boundary are then classified as anomalies.

Second, Isolation Forest (IF) leverages an ensemble of isolation trees to detect anomalies \cite{liu2012}. Each tree is constructed by recursively partitioning the data based on randomly selected features and split points. The algorithm aggregates these path lengths to provide an anomaly score for each data point.

Finally, Deep Support Vector Data Description (D-SVDD) utilizes a deep neural network to learn a hypersphere in the feature space that encapsulates the training data while minimizing the volume of this hypersphere \cite{ruff2018}.

Given the potential for variation in the rankings produced by each classifier, we introduce an ensemble algorithm to harmonize their predictions and enhance the reliability of the utterance selection process (see \cref{alg:extract}). This algorithm prioritizes consistently flagged utterances similar to those of all three classifiers, ensuring a more robust and confident data selection for augmenting the SSL model's training set.

\cref{alg:extract} aims to create a list of utterances ($R$), given three pre-sorted lists ($U_1$, $U_2$, $U_3$) based on the scores of Deep SVDD, One-class SVM, and Isolation Forest. The algorithm prioritizes utterances appearing high in all three lists.

The process begins by initializing $L$ with an initial ranking limit ($L_0$). It iteratively expands $L$ until the total duration of utterances in $R$ meets a predefined time limit ($k$ hours). Each iteration involves selecting top $L$ utterances from each list ($U_1$, $U_2$, $U_3$), denoted as $\widehat{U_1}$, $\widehat{U_2}$, and $\widehat{U_3}$. Subsequently, each utterance ($u$) in $\widehat{U_1}$ is evaluated. If $u$ is not already in $R$ and is present in both $\widehat{U_2}$ and $\widehat{U_3}$, it signifies high ranking across all three lists and is appended to $R$. After processing all utterances in $\widehat{U_1}$, $L$ is incremented by $L_0$ for the next iteration. This cycle continues until the time constraint of $R$ is met.

\section{Experiments}
\label{sec:experiments}

Our experimental procedure unfolds as follows: In addition to leveraging the existing, limited volume of target language data, we employ the top-$k$ hours of sampled data from the ML-SUPERB multilingual corpus. This selection is facilitated by a one-class classifier or \cref{alg:extract}, intended for the continued pre-training of the SSL model. The refined model is then utilized to generate SSL features, enabling fine-tuning of the downstream ASR model using target language speech data accompanied by transcriptions. For our framework, Fairseq is used for pre-training, while both S3PRL and ESPnet are employed for downstream ASR tasks \cite{ott2019,watanabe2018,yang2021}.

\subsection{Continued Pre-training}
\label{ssec:continued_pre_training}

Following the methodology outlined by Nowakowski et al., we continued to pre-train the entire SSL model with the curated data. The training utterances were curated from the ML-SUPERB benchmark using one-class classifiers. This process involved $100,000$ updates, including $10,000$ warmup steps, with a batch size of $256$ and a learning rate set at $10^{-4}$. All settings were aligned with those of the \texttt{wav2vec 2.0 large} model. The optimal configuration was identified using validation data from the ML-SUPERB benchmark.

\subsection{One-class Classification}
\label{ssec:one_class_clf}

Our process commenced using NVIDIA's pre-trained TitaNet-LID to derive language embeddings for each utterance \cite{jia2023}. This model attains a $7.0$\% error rate on the $103$ languages test set. The obtained embeddings are situated in a 512-dimensional space, originating from the penultimate layer.

\begin{table*}[t]
\centering
\scalebox{0.9}{
\begin{tabular}{lcccccc}
\toprule
\textbf{SSL/Acoustic Features} & \textbf{}{\# Params. (M)} & \textbf{\# Languages} & \textbf{Amis-10min} & \textbf{Amis-1h} & \textbf{Seediq-10min} & \textbf{Seediq-1h}\\
\toprule
Mel-filterbank & N/A & N/A & 40.5 & 28.5 & 49.4 & 36.0 \\
\midrule
wav2vec2-base \cite{baevski2020} & 95 & 1 & 14.6 & 9.8 & 19.8 & 11.9 \\
wav2vec2-large \cite{baevski2020} & 317 & 1 & 13.4 & 9.5 & 18.5 & 11.4 \\
robust-wav2vec2-large \cite{hsu2021} & 317 & 1 & 13.5 & 9.3 & 18.6 & 11.5 \\
\midrule
chinese-wav2vec2-base\footnotemark \cite{zhang2022} & 95 & 1 & 14.4 & 9.4 & 18.7 & 11.2 \\
chinese-wav2vec2-large\footnotemark \cite{zhang2022} & 317 & 1 & 89.4 & 10.0 & 22.2 & 12.5 \\
chinese-wav2vec2-large$^{*}$ \cite{zhang2022} & 317 & 1 & 15.3 & 10.0 & 21.6 & 12.3 \\
\midrule
wav2vec2-base-23 \cite{wang2021} & 95 & 23 & 14.5 & 9.2 & 19.6 & 12.1 \\
wav2vec2-large-23 \cite{wang2021} & 317 & 23 & 103.4 & 9.6 & 19.2 & 11.3 \\
wav2vec2-large-23$^{*}$ \cite{wang2021} & 317 & 23 & 13.5 & 9.2 & 18.4 & 11.4 \\
\midrule
XLSR-53 \cite{conneau2020} & 317 & 53 & 81.4 & 10.4 & 18.6 & 14.0 \\
XLSR-300M\footnotemark \cite{babu2022} & 317 & 128 & 90.8 & 7.7 & 14.0 & 8.4 \\
XLSR-300M$^{*}$ \cite{babu2022} & 317 & 128 & 10.9 & 7.9 & 13.7 & 8.6 \\
\midrule
HuBERT-base \cite{hsu2021a} & 95 & 1 & 12.6 & 8.8 & 18.2 & 11.0 \\
HuBERT-large \cite{hsu2021a} & 317 & 1 & 80.6 & 7.4 & 84.1 & 8.9 \\
HuBERT-large$^{*}$ \cite{hsu2021a} & 317 & 1 & 10.3 & 7.5 & 14.0 & 9.1 \\
\midrule
chinese-hubert-base\footnotemark \cite{zhang2022} & 317 & 1 & 13.2 & 9.0 & 18.4 & 10.9 \\
chinese-hubert-large\footnotemark \cite{zhang2022} & 317 & 1 & 11.0 & 7.3 & 14.7 & 8.1 \\
\midrule
mHuBERT-base \cite{lee2022} & 95 & 3 & 12.9 & 8.9 & 17.3 & 11.5 \\
\bottomrule
\end{tabular}
}
\caption{CER (\%) evaluated on Amis and Seediq using SSL models. $^{*}$ indicates a smaller downstream ASR model. 10min and 1h mean the amount of training data used in downstream ASR models.}
\label{tab:experiment_results}
\vspace{-15pt}
\end{table*}

\footnotetext[3]{https://huggingface.co/TencentGameMate/chinese-wav2vec2-base}
\footnotetext[4]{https://huggingface.co/TencentGameMate/chinese-wav2vec2-large}
\footnotetext[5]{https://huggingface.co/facebook/wav2vec2-xls-r-300m}
\footnotetext[6]{https://huggingface.co/TencentGameMate/chinese-hubert-base}
\footnotetext[7]{https://huggingface.co/TencentGameMate/chinese-hubert-large}

For the classifiers, we utilized scikit-learn to train the One-class SVM and Isolation Forest, largely sticking to the default parameters. In the case of Deep SVDD, we implemented a modification by substituting the original 2D convolution for a 1D variant. The autoencoder was pre-trained for $2,500$ epochs with a learning rate of $10^{-2}$, and the encoder was further trained for $1,000$ epochs at a reduced learning rate of $10^{-3}$. Negative samples were selected from the ML-SUPERB training data and combined with the original test utterances from Amis and Seediq as positive samples. Our experiments demonstrated the superior performance of Deep SVDD in this setup.

\subsection{Fine-tuning}
\label{ssec:fine-tuning}

The downstream ASR task followed ML-SUPERB's design. The frozen SSL representations were weighted summed with learnable weights. The ASR model consisted of a CNN layer that downsamples the SSL feature sequence to half the original length and two Transformer encoders with an attention dimension of $256$, a feedforward layer with a dimension of $1,024$, and $8$ heads. The model is trained with connectionist temporal classification (CTC) and Adam optimizer with a learning rate of $10^{-4}$ and $10^{-6}$ weight decay \cite{kingma2015,graves2006}. We set dropout to $0.1$, batch size to $8$, and gradient accumulation to $4$. SpecAugment was applied to SSL features \cite{park2019}.

\section{Results}
\label{sec:resulrs}


\subsection{ASR Results on Amis and Seediq}
\label{ssec:asr_result}

\begin{figure*}[t]
\centering
\begin{subfloat}{
    \includegraphics[width=0.45\textwidth]{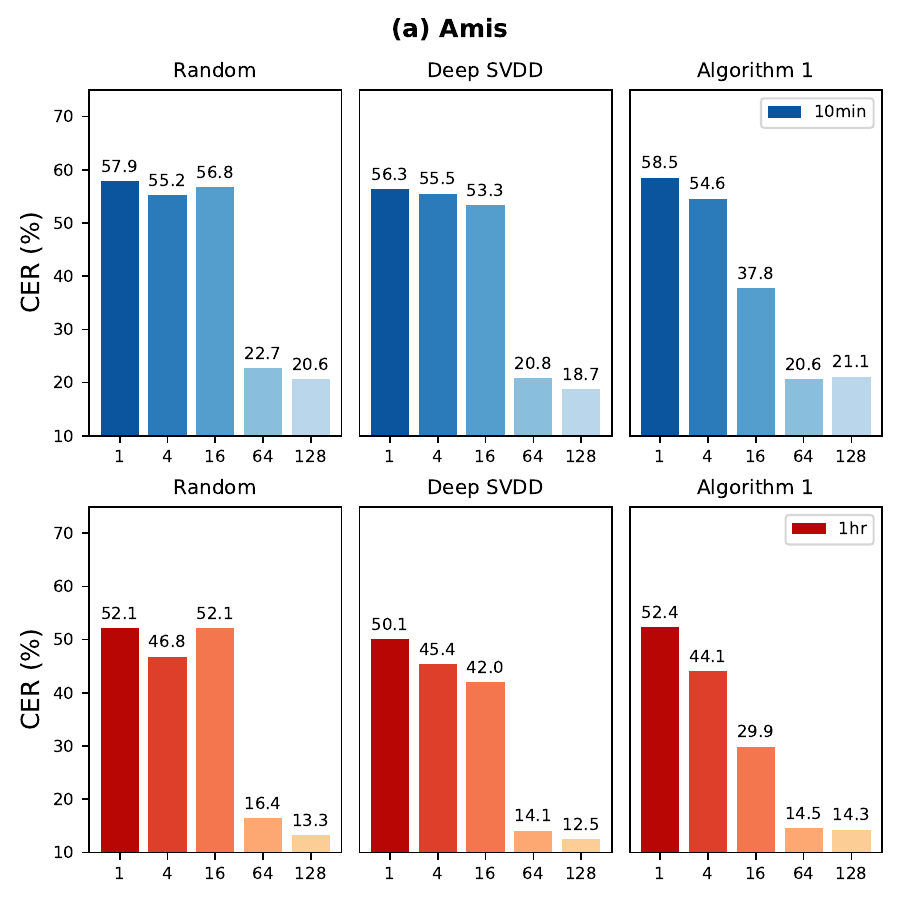}
}
\end{subfloat}
\hfill
\begin{subfloat} {
\includegraphics[width=0.45\textwidth]{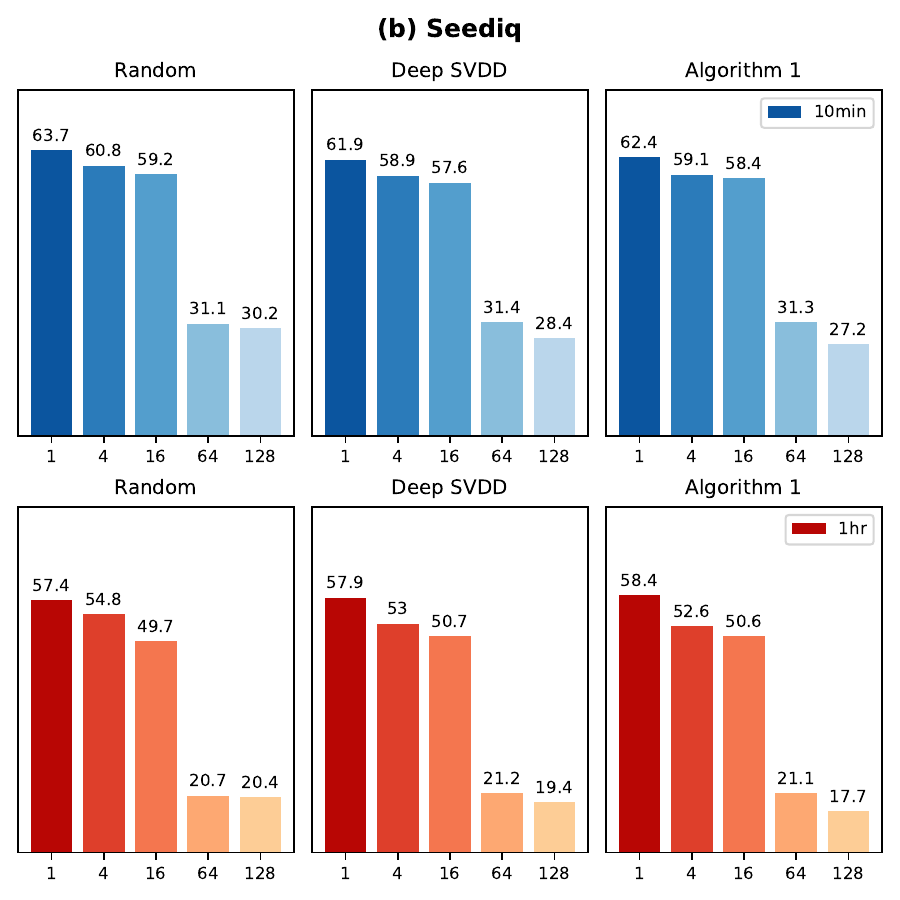}
}
\end{subfloat}

\caption{The effect of data amount and selection on continued pre-training XLSR-128 for Amis and Seediq. The x-axis represents the amount of sampled data (in hours), and `1' means 1 hour of non-target language data is used together with 1 hour of target language data in continued pre-training. The shades in blue and red represent the amount of pre-training data.}
\label{fig:results}
\vspace{-15pt}
\end{figure*}

\cref{tab:experiment_results} presents an overview of ASR results for both Amis and Seediq, where the downstream ASR models are directly fine-tuned from SSL models in various settings (see \cref{ssec:fine-tuning} for details). We have compared models that were pre-trained with utterances in multilingual, English-only, and Mandarin Chinese-only. Additionally, we evaluated the effectiveness of the model architecture and parameter size.

\texttt{chinese-hubert-large} (Hubert-large fine-tuned with Mandarin Chinese data) consistently outperformed other models across different training data volumes, except for the Seediq 10-minute task. This superior performance was observed despite using identical training data for \texttt{wav2vec2} (see results of chinese-hubert-large vs. chinese-wav2vec2-large), suggesting potential advantages of the \texttt{HuBERT} architecture for these low-resource scenarios. Furthermore, \texttt{XLSR-300M} demonstrated comparable effectiveness, achieving the lowest CER of 13.7\% on the Seediq 10-minute task. These results highlight several noteworthy trends:

\begin{itemize}[left=0pt,noitemsep]
    \item \texttt{HuBERT} generally outperform \texttt{wav2vec2} when trained on the same data.
    \item Larger SSL models yield modest but consistent gains, especially in the 1-hour training setup.
    \item Multilingual pre-training contributes positively to ASR performance in low-resource settings.
\end{itemize}

As observed in \cref{tab:experiment_results}, large SSL models, while generally outperforming their smaller counterparts, exhibit a tendency to overfit when trained on limited data. This overfitting was particularly evident for models like \texttt{wav2vec2-large-23}, \texttt{chinese-wav2vec2-large}, \texttt{XLSR-53}, \texttt{XLSR-300M}, and\\ \texttt{HuBERT-large}. To mitigate this issue, we modified the downstream ASR model architecture by halving the number of attention heads to $4$, reducing the feedforward dimension to $512$, and increasing the dropout rate to $0.3$. These adjustments, as demonstrated in \cref{tab:experiment_results} (models denoted by $^{*}$), enabled the smaller ASR model to achieve comparable performance even when paired with larger SSL models, effectively addressing the overfitting problem.

\subsection{Proposed Data Augmentation Method}
\label{ssec:data_augmentation}

This section explores the effectiveness of our proposed data augmentation method. Specifically, we attempt to improve the downstream ASR results that we discussed previously (see \cref{ssec:asr_result}) by leveraging our proposed data augmentation method (see \cref{sec:research} for more details).

While \texttt{XLSR-300M} did not consistently achieve the highest performance across all experimental settings, we selected it as our primary benchmark model. This decision stems from the fact that \texttt{XLSR-300M} is a multilingual model not specifically pre-trained on languages closely related to Amis and Seediq, making it a more generalizable choice for broader low-resource language research where such closely related pre-trained models might not be readily available.

\cref{fig:results} illustrates the impact of data volume and selection strategy on the performance of \texttt{XLSR-300M} after continued pre-training for both Amis and Seediq. We evaluated three data selection approaches: random sampling, Deep SVDD, and our proposed ensemble algorithm (\cref{alg:extract}). As expected, increasing the amount of pre-training data generally leads to improved ASR performance, mitigating the risk of overfitting. Notably, using only $1$ hour of data from ML-SUPERB resulted in suboptimal performance across all selection methods, underscoring the importance of sufficient training data volume. Expanding the pre-training data to $64$ hours significantly reduced character error rate (CER) for both languages, highlighting the consistent benefit of larger datasets across languages.

Our proposed methods (SVDD and \cref{alg:extract}) consistently outperform the random selection baseline across nearly all experiments. The only exception is noted in the Seediq 1-hour experiments, where our methods demonstrate their effectiveness after selecting $128$ hours of non-target language data. Specifically, in both the Seediq 10-minute and 1-hour scenarios with $128$ hours of data selection, \cref{alg:extract} consistently yields superior results compared to both random selection and SVDD. In contrast, SVDD performs the best in Amis 10-minute and 1-hour scenarios with $128$ hours of data selection. However, increasing the data size to $128$ hours yielded only marginal improvements, suggesting that the model might be approaching saturation given the limited representation of the target languages within the available multilingual speech data.

Our findings for Amis reveal that when pre-training data is limited (e.g., $16$ hours), utilizing \cref{alg:extract}) for data selection significantly outperforms both random sampling and Deep SVDD, particularly in the 10-minute and 1-hour training scenarios. Moreover, \cref{fig:results} demonstrates that the SSL model pre-trained with data selected using our algorithm exhibits greater robustness across almost all experimental conditions than models trained with randomly selected or Deep SVDD-sampled data.
\section{Conclusions}
\label{sec:conclusions}

This study introduces speech corpora for two low-resource Austronesian languages, Amis and Seediq, and presents a data augmentation strategy for enhancing low-resource ASR. We propose leveraging a novel ensemble algorithm to select acoustically similar utterances from a multilingual corpus for continued pre-training of the \texttt{XLSR-300M} model. Our results demonstrate that while the performance improvements are currently modest, the positive correlation between pre-training data volume and ASR accuracy highlights the potential of this approach.

\clearpage

\printbibliography[heading=bibnumbered]


\end{document}